\documentclass[letterpaper, 10pt, conference]{ieeeconf}
\usepackage{amsmath,amsfonts}
\usepackage{algorithmic}
\usepackage{algorithm}
\usepackage{array}
\usepackage[font=small]{caption}
\usepackage[font=small]{subcaption}
\usepackage{graphicx}
\IEEEoverridecommandlockouts
\newcommand{\refframe}[1]{\mathcal{#1}}
\renewcommand{\vec}[1]{\boldsymbol{#1}}

\newcommand{\mat}[1]{\mathbf{#1}}
\newcommand{\tpose}{\mathsf{T}}
\newcommand{\vecframe}[2]{\vec{#1}^{\refframe{#2}}}
\newcommand{\rotation}[2]{\mat{R}_{\refframe{#1}}^{\refframe{#2}}}

\DeclareMathOperator{\diag}{diag}

\begin{document}

\title{\LARGE \bf Imitation learning for sim-to-real adaptation of robotic cutting policies based on residual Gaussian process disturbance force model}

\author{Jamie Hathaway$^{1,2,\dagger}$, Rustam Stolkin$^{1,2}$, Alireza Rastegarpanah$^{1,2, \dagger}$%
\thanks{1 Department of Metallurgy \& Materials Science, University of Birmingham, Birmingham, UK, B15 2TT.}%
\thanks{2 The Faraday Institution, Quad One, Harwell Science and Innovation Campus, Didcot, UK, OX11 0RA.}%
\thanks{$\dagger$ These authors contributed equally to this work.}%
\thanks{This work was supported by the project ``Research and Development of a Highly Automated and Safe Streamlined Process for Increase Lithium-ion Battery Repurposing and Recycling'' (REBELION) under Grant 101104241}
\thanks{The authors would like to acknowledge Abdelaziz Wasfy Shaarawy, Carl Meggs and Christopher Gell respectively for assistance with experimental validation, design of material holder and cutter tool for experiments herein.}
\thanks{© 2024 IEEE.  Personal use of this material is permitted.  Permission from IEEE must be obtained for all other uses, in any current or future media, including reprinting/republishing this material for advertising or promotional purposes, creating new collective works, for resale or redistribution to servers or lists, or reuse of any copyrighted component of this work in other works.}}

\markboth{IEEE/RSJ International Conference on Intelligent Robots and Systems}{Hathaway et al.: Sim-to-real adaptation of robotic cutting}

\maketitle

\begin{abstract}
Robotic cutting, a crucial task in applications such as disassembly and decommissioning, faces challenges due to uncertainties in real-world environments. This paper presents a novel approach to enhance sim-to-real transfer of robotic cutting policies, leveraging a hybrid method integrating Gaussian process (GP) regression to model disturbance forces encountered during cutting tasks. By learning from a limited number of real-world trials, our method captures residual process dynamics, enabling effective adaptation to diverse materials without the need for fine-tuning on physical robots. Key to our approach is the utilisation of imitation learning, where expert actions in the uncorrected simulation are paired with GP-corrected observations. This pairing aligns action distributions between simulated and real-world domains, facilitating robust policy transfer. We illustrate the efficacy of our method through real world cutting trials in autonomously adapting to diverse material properties; our method surpasses re-training, while providing similar benefits to fine-tuning in real-world cutting scenarios. Notably, policies transferred using our approach exhibit enhanced resilience to noise and disturbances, while maintaining fidelity to expert behaviours from the source domain.
\end{abstract}

\section{Introduction}

Contact cutting processes feature extensively in a range of applications, most notably manufacturing. However, robotic disassembly applications often struggle with variable product designs, condition uncertainties, and the absence of critical manufacturer information, such as material knowledge for process parameter selection or CAD models for process planning. Hence, for such applications, it is desirable to rapidly adapt to novel products with minimal prior knowledge about the components or fasteners to be cut. With the aid of simulation environments, learning-based approaches have shown promise in adapting to uncertainties across various applications, albeit mainly to non-destructive tasks \cite{HybridTrajectoryForceLearningAssembly, PegInHoleDDPGVIC, SimToRealDomainRandomisationPushingTask}, though sim-to-real adaptation for destructive tasks like cutting or milling, although possible \cite{LearningRoboticMillingRL}, remains challenging.

In the context of robotic cutting, besides safety issues and limited availability of labelled target domain data, notable challenges in sim-to-real adaptation include sensor limitations such as noise, disturbances \cite{IdentificationDisturbanceObserver} and residual unmodelled dynamic effects such as chattering, which are challenging to model without laborious identification. Learning-based approaches suffer poor generalisation when dealing with distributional mismatch between source and target domain observations. Furthermore, direct re-training is problematic due to the problem of catastrophic forgetting, resulting in large reductions in performance \cite{ContinualDeepLearningTimeSeriesModelling}. Previous approaches to sim to real adaptation of can be broadly categorised into domain adaptive and transfer learning approaches \cite{SimToRealRLReview}. Various domain adaptive approaches have been proposed based on classifier and discriminator models \cite{ADATimeDomainAdaptation}, and for tool wear classification in milling \cite{AdversarialDomainAdaptation}. To surmount the problem of limited target domain data, generative approaches have also been proposed based on synthesis of target domain data \cite{GenerativeNNBasedDomainAdaptationIncompleteTargetDomain}, \cite{OneShotDomainAdaptiveImitationLearning}. For generative and discriminator-based approaches, in addition to labelled source domain data, a large amount of \emph{unlabelled} target domain data are necessary to train the discriminators. Other approaches have been also proposed based on translation models \cite{UnpairedImage2ImageRL} and learning of unified feature representations between domains \cite{DeepSphericalManifoldGPDomainAdaptation}, \cite{RLMachiningDeformationControl}, \cite{DomainAdaptationRLUnifiedLatentRepresentation}. A common assumption is that the conditional distributions of outputs, such as classifiers, are domain invariant, while differences between the domains arise from differences in the marginal distributions over observations (covariate shift), which is not always the case \cite{DomainAdaptationTargetConditionalShift}. In this work, we consider the case that the conditional distributions differ between domains, representing the ``conditional shift'' case from \cite{DomainAdaptationTargetConditionalShift}. Other approaches aim to directly compensate measured disturbances on the real setup, as proposed in \cite{GPDisturbanceObserver}, based on a combination of observer with a Gaussian process (GP) model of position-based disturbances. More specifically to milling, \cite{IdentificationDisturbanceObserver} proposed a 4-inertial disturbance observer approach for identification of milling force models. However, the authors suggest that some level of prior knowledge may be required to obtain initial estimates for accurate parameter identification.

\begin{figure*}[t]
    \centering
    \captionsetup{aboveskip=10pt,belowskip=-15pt}
    \includegraphics[width=0.73\textwidth]{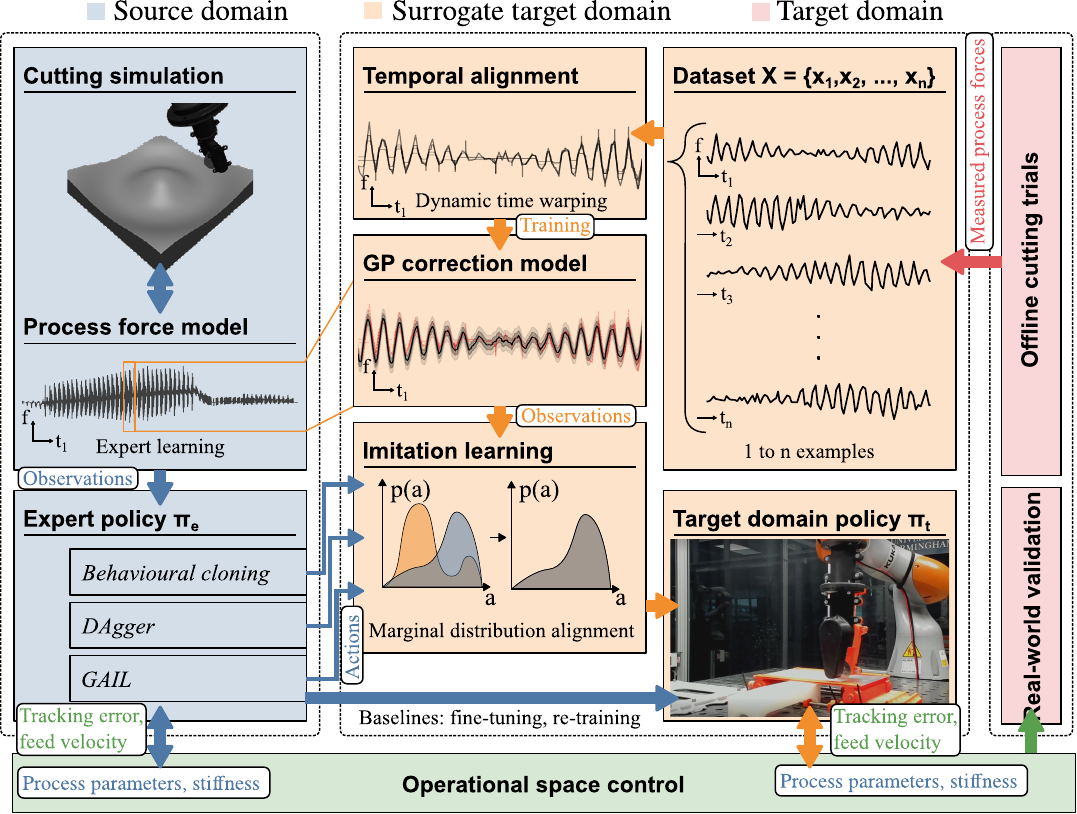}
    \caption{Overview of the proposed framework, consisting of three stages. In the first stage, a model of the cutting mechanics (source domain) is employed to train an expert policy. Secondly, cutting process force data are collected offline on a target domain (real world), which is used to train a corrective model of disturbances during the real world cutting process. Finally, imitation learning on a surrogate target domain is employed to align the marginal action distributions of expert and learner policies to generate a new target domain policy.}
    \label{fig:Method-Framework-Overview}
\end{figure*}

Transfer learning approaches contrast domain adaptive approaches in that they aim to improve performance on a target domain via uni- or bi-directional transfer of knowledge between domains. Recently, sim-to-real transfer approaches have been proposed based on augmentation of simulations with models learned from trajectories collected from the real environment \cite{SimToRealNeuralAugmentedSimulation}, \cite{DeepInverseDynamicModelSim2Real}. However, these approaches require a large number of real world samples to train. On the other hand, GPs have been demonstrated in other application areas to be efficient at learning from a small number of samples \cite{SimToRealBasedOnGPandDR}, \cite{GPDomainAdaptationMultipleExperts}. Alternative sim-to-real transfer approaches have employed system identification methods to optimise physical parameters of a simulation to maximise target domain performance \cite{Sim2RealRLWithoutDR}. Therefore, these methods are capable of only modelling parametric differences between source and target domains. Related to this concept and this present work, \cite{OfflineImitationLearningMisspecifiedSim} leverages an imitation-learning-based approach for domain adaptation based on real world demonstrations on a misspecified simulator.

This paper aims to address challenges in sim-to-real transfer of cutting by learning a GP model of residual process dynamics from minimal ($<$20) real world demonstrations. To address the problem of changing decision boundaries for actions between domains, we propose an approach based on imitation learning over corrected simulation observations to align the marginal action distributions between source and target domains. We demonstrate the proposed method outperforms direct application of the expert on the target domain. Our approach provides similar benefits to fine-tuning, however, results in policies that are generally more robust to noise and disturbances and correspond more closely to the source domain expert policy behaviour. An overview of our framework is shown in Figure \ref{fig:Method-Framework-Overview}.

\section{Methodology\label{sec:Method}}

\subsection{Dataset Preparation}

We consider an initially unstructured dataset comprising time series $\mathcal{D}_{u} = \{\mat{X}_{0}\ldots,\mat{X}_{n}\}$. Each times series contains force measurements $\mat{X}_{i} = \{(t_{i}, \vec{f_{e}})\}$. Many periodic disturbances are parameterised as a function of position instead of time \cite{GPDisturbanceObserver}. In the absence of position measurements, the data must first be aligned in the time domain. Each time series is first normalised as:
\begin{equation}
    \mat{X}_{i} = \Sigma_{\mat{X}_{i}}^{-1}\left(\mat{X}_{i} - \bar{\mat{X}}_{i}\right)
\end{equation}
where $\bar{\mat{X}}_{i}$, $\Sigma_{\mat{X}_{i}}$ are the mean and standard deviations of each example time series $\mat{X}_{i}$. The alignment in time domain consists of an initial coarse alignment stage, followed by a fine alignment stage. The coarse alignment is performed by maximising the cross-correlation:
\begin{equation}
    t_{\mathrm{delay}} = \underset{t}{\mathrm{argmax}}\left((\mat{X}_{i}*\mat{X}_{j}))(t)\right)
\end{equation}

Subsequently, the fine alignment is performed with pairwise dynamic time warping (DTW) between each example trajectory pair $\mat{X}_{i}$, $\mat{X}_{j}$. DTW finds an optimal warping path:
\begin{equation}
   (l^{*}_{n}, m^{*}_{n}) = \underset{(l_{n},m_{n})}{\mathrm{argmin}} \sum_{n} \frac{d(\vec{x}_{l_{n}}, \vec{x}_{m_{n}}) w_{n}}{\sum_{n^{\prime}}{w_{n^{\prime}}}}\quad
   \label{eq:DTW-Problem}
\end{equation}
with respect to a cumulative weighted distance between features $d(\vec{x}_{l}, \vec{x}_{m})$ (here Euclidean distance), where $w_{n}$ are weighting terms. Adopting the convention from \cite{DTWPackageR}, the ``symmetric2'' step pattern was used. To account for variable length examples, we employ an open-ended DTW approach, which allows for more flexible alignment by permitting variable-length warping paths, which further computes the minimum \eqref{eq:DTW-Problem} over all truncated sequences of $\mat{X}_{j}$. To re-index a time series according to the warping path, each element of the original time series is assigned to a new time point based on the optimal mapping. This re-indexed time series $\hat{\mat{X}}$ effectively accounts for time distortions, allowing synchronization and alignment of the data across different examples. Hence, the final dataset comprises aligned time series $\mathcal{D} = \{\hat{\mat{X}}_{0}\ldots,\hat{\mat{X}}_{n}\}$.

\subsection{GP Force Model}

We consider the regression problem:
\begin{equation}
    \vec{f}_{e} = \vec{f}(t) + \vec{d}(t) + \epsilon
\end{equation}
where $f(t)$ represents the force computed from the underlying process force model, while $\vec{d}(t)$ represents a disturbance force. The disturbance $\vec{d}(t)$ is assumed to be periodic and with independent and identically distributed (i.i.d) noise $\epsilon\sim\mathcal{N}(0, \sigma^{2})$. We model the disturbance force $\vec{d}(t)$ as a Gaussian Process (GP) with zero mean and covariance function $k(\vec{x}, \vec{x}^{\prime})$
\begin{equation}
    \vec{d}(\vec{t}) \sim \mathcal{GP}(\vec{0}, k(\vec{t}, \vec{t}^{\prime}))
\end{equation}
where the distribution of observed and unobserved data is modeled as a joint multivariate Gaussian distribution:
\begin{equation}
    \begin{bmatrix}
        \vec{d} \\
        \vec{d}_{*}
    \end{bmatrix} = \mathcal{N}\left(\mat{0},
    \begin{bmatrix}
        \mat{K} & \mat{K}_{*} \\
        \mat{K}_{*}^{\tpose} & \mat{K}_{**}
    \end{bmatrix}\right)
\end{equation}
where $\mat{K}$, $\mat{K}_{*}$, $\mat{K}_{**}$ are the training, train-test and test covariance matrices respectively. The posterior predictive distribution for the disturbance force $\vec{d}_{*}$ given test points $\mat{X}_{*}$ is then
\begin{align}
p(\vec{d}_{*}|\mathcal{D},\vec{d},\mat{X}_{*}) = & \mathcal{N}\left(\vec{\mu},\mat{\Sigma}\right)\label{eq:Posterior-Distribution} \\
    \vec{\mu} = & \mat{K}_{*}^{\mathsf{T}}\left[\mat{K}+\sigma^{2}\mat{I}\right]^{-1}\vec{d} \\
    \mat{\Sigma} = & \mat{K}_{**}-\mat{K}_{*}^{\mathsf{T}}\left[\mat{K}+\sigma^{2}\mat{I}\right]^{-1}\mat{K}_{*}
\end{align}
To capture the periodic nature of the disturbance force, we use the exponential sine covariance function
\begin{equation}
    k(\vec{x}, \vec{x}^{\prime}) = \exp\left(\frac{2\sin^2\left(\frac{\pi d(\vec{t}, \vec{t}^{\prime})}{p}\right)}{l^{2}} \right)
\end{equation}
with periodicity $p$, length scale $l$ and Euclidean distance $d(\vec{t}, \vec{t}^{\prime})$. The kernel hyperparameters for the GP model are estimated directly from the data by minimising the negative (marginal) log likelihood
\begin{multline}
\log p(\vec{d}|\mat{X})=-\frac{1}{2}\vec{d}^{\tpose}\left[\mat{K}+\sigma^{2}\mat{I}\right]^{-1}\vec{d}\\-\frac{1}{2}\log\left|\mat{K}+\sigma^{2}\mat{I}\right|-\frac{n}{2}\log(2\pi)
\end{multline}
over the hyperparameter space. To make this procedure more robust, we provide some initial estimates for the noise level based on the sensor noise (measured under static loading in free space) and perform the optimisation from different initialisations.

Based on the assumptions of the mechanistic force modelling approach \cite{TheMachiningOfMetals}, the cutting force $\vec{f}$ can be related to each cutting edge $p\in{1\ldots N_{p}}$ of a fluted cutting tool via mechanistic constants $\vec{k}_{c}$, $\vec{k}_{e}$:
\begin{equation}
	\vecframe{f}{P}=b_{p}\vec{k}_{e} + b_{p}\vec{k}_{c}h_{p}
	\label{eq:Method-Flute-Force}
\end{equation}
where $b_{p}$ is the thickness of the cutting edge, $h_{p}$ is the uncut chip thickness, computed from the cutting edge rotation angle $\theta_{p}$ in the tool model reference frame, tool feed rate $v$ and spindle speed $\omega$: 
\begin{equation}
	h_{p}= \sin\theta_{p}\frac{v}{N_{p}\omega}
\end{equation}
The total cutting force is computed as the sum over cutting elements weighted by the Boolean vector $\mat{G}\in\mathbb{B}^{N_{p}}$ specifying the engagement of each element $k$ with the workpiece.
\begin{equation}
	\vec{f}=\sum_{p}^{N_{p}} G_{p}\rotation{P}{}\vecframe{f}{P}
	\label{eq:Method-Mechanistic-Total-Force}
\end{equation}
where $\rotation{P}{}$ is the rotation of cutting edge $p$ about the tool axis to the tool model frame $M$.

\subsection{Imitation Learning Framework}
The cutting task is formulated in the Mujoco simulation environment, into which the model of cutting mechanics from \eqref{eq:Method-Mechanistic-Total-Force} is embedded. The expert reward function $r$ expresses a weighted sum (by weights $Q_{\boldsymbol{\cdot}}$) of task-specific performance metrics $\mathrm{MRV}$ (material removed volume) and cutting time $t_\mathrm{cut}$, and feasibility reward shaping comprising path error $\vec{e}$, process force $\vec{f}$
\begin{equation}
r = Q_{\textrm{MRV}}\cdot\mathrm{MRV} - Q_{\mathrm{cut}}t_{\mathrm{cut}} - \vec{e}\mat{Q}_{d}\vec{e}^\tpose - \vec{f}\mat{Q}_{f}\vec{f}^\tpose\label{eq:RL-Reward-Function}
\end{equation}
For the presented cutting task, the actions of the agent considered are $\vec{a}=\begin{bmatrix}\diag^{-1}\left({\mat{K}_{p}}\right) & \dot{t}_{\Delta} & \dot{n_{\Delta}} \end{bmatrix}^{\tpose}$, where $\dot{t}_{\Delta}$ is related to the commanded feed rate as $t_{\Delta}=\frac{v}{v_{n}} - 1$, where $v_{n}$ is a nominal feed rate, and $n_{\Delta}$ is the commanded depth of cut (DoC - here radial depth of cut). 
The observation vector was defined as $\vec{o} = \begin{bmatrix}\dot{\vec{c}}^{\tpose}(t)\dot{\vec{v}}_{EE} & \vec{e} & \vec{v}_{EE} & \vec{f}_{e} & t_{\Delta} & n_{\Delta} & \diag^{-1}\left({\mat{K}_{p}}\right) \end{bmatrix}^{\tpose}$, where $\vec{c}(t)$ is the reference cutting path, $\vec{v}_{EE}$ the end-effector velocity. The reward function weights and hyperparameters were selected as our previous work \cite{LearningRoboticMillingRL}. Due to the problems of catastrophic forgetting with fine-tuning and high cost of data collection in the real environment, we propose an imitation-learning based approach to train a target policy that can adapt to the target domain via a surrogate target domain. We establish a test case comprising ``offline'' and ``online'' imitation learning algorithms; behavioural cloning (BC) and DAgger, which we contrast with the case of the expert policy directly transferred with no fine-tuning, and the expert with fine-tuning directly on the surrogate target domain. 

During training, we employ the source domain expert policy $\pi_{e}$ to train a surrogate target domain policy $\hat{\pi}_{t}$. To resemble the fine-tuning case, we initialise the target domain policy as $\hat{\pi}_{t}=\pi_{e}$. At each step, the tuple $(\vec{o}_{e}, \pi_{e}(\vec{o}_{e}))$ is sampled from the base environment using the data collection policy 
\begin{equation}
    \pi_{d} = \beta\pi_{e} + (1-\beta)\pi_{t}
\end{equation}
where $\beta$ is the non-expert action probability, which is non-zero for DAgger and zero otherwise. Then, the expert observations $\vec{o}_{e}$, actions $\vec{a}$ $(\vec{o}_{e}, \vec{a})$ are modified by sampling from the posterior distribution \eqref{eq:Posterior-Distribution} for each point in the trajectory to generate new experiences $(\vec{o}_{t}, \vec{a})$. Subsequently, the surrogate target domain experiences are used to update the target policy as the standard behavioural cloning procedure. However, the method is applicable to other imitation learning algorithms such as GAIL or AIRL. This procedure is summarised in Algorithm \ref{alg:Method-ILGP-Framework}. We employ the proximal policy optimisation (PPO) algorithm for learning and fine-tuning of policies, although the principles are independent of learning algorithm. For each approach, we conduct training with all strategies for 50 episodes, a learning rate of $1\times 10^{-3}$ and batch size of 64. For DAgger, $\beta$ is varied according to a 0--1 linear schedule for 45 episodes.
\vspace{-8pt}

\begin{algorithm}

    \caption{Imitation-learning sim-to-real transfer}
    \label{alg:Method-ILGP-Framework}
    \begin{algorithmic}
    \STATE Expert policy, $\pi_{e}$, target policy $\hat{\pi}_{t}$
    \STATE Source domain (expert) environment $\mathcal{E}$
    \STATE Disturbance model $p(\vec{d}^{\prime}|\mathcal{D},\vec{x},\vec{d},\vec{x}^{\prime})$
    \STATE $\hat{\pi}_{t} \gets \pi_{e}$
    \FOR {$i = 0$ \TO $N$}
        \WHILE{episode \NOT done}
        \STATE Sample $(\vec{o}_{e}, \pi_{e}(\vec{o}_{e}))$ from $\mathcal{E}$
        \STATE Sample $\vec{d}^{\prime} \sim p(\vec{d}^{\prime}|\mathcal{D},\vec{x},\vec{d},\vec{x}^{\prime})$
        \STATE $\vec{o}_{t} \gets \vec{o}_{e} + \vec{d}^{\prime}$
        \STATE Append $\mathcal{D}_{e}$ with $(\vec{o}_{t}, \pi_{e}(\vec{o}_{e}))$
        \STATE Update $\mathcal{E}$ with $a \sim \hat{\pi}_{t}(\vec{o}_{e})$
        \ENDWHILE
    \ENDFOR
    \STATE Train $\hat{\pi}_{t}$ with $\mathcal{D}_{e}$ as $\mathsf{BC}, \mathsf{DAgger}, \ldots$
    \end{algorithmic}
\end{algorithm}
\vspace{-15pt}

\subsection{Experimental Setup}

\begin{figure}
    \centering
    \captionsetup{aboveskip=6pt,belowskip=-20pt}
    \includegraphics[width=0.7\columnwidth]{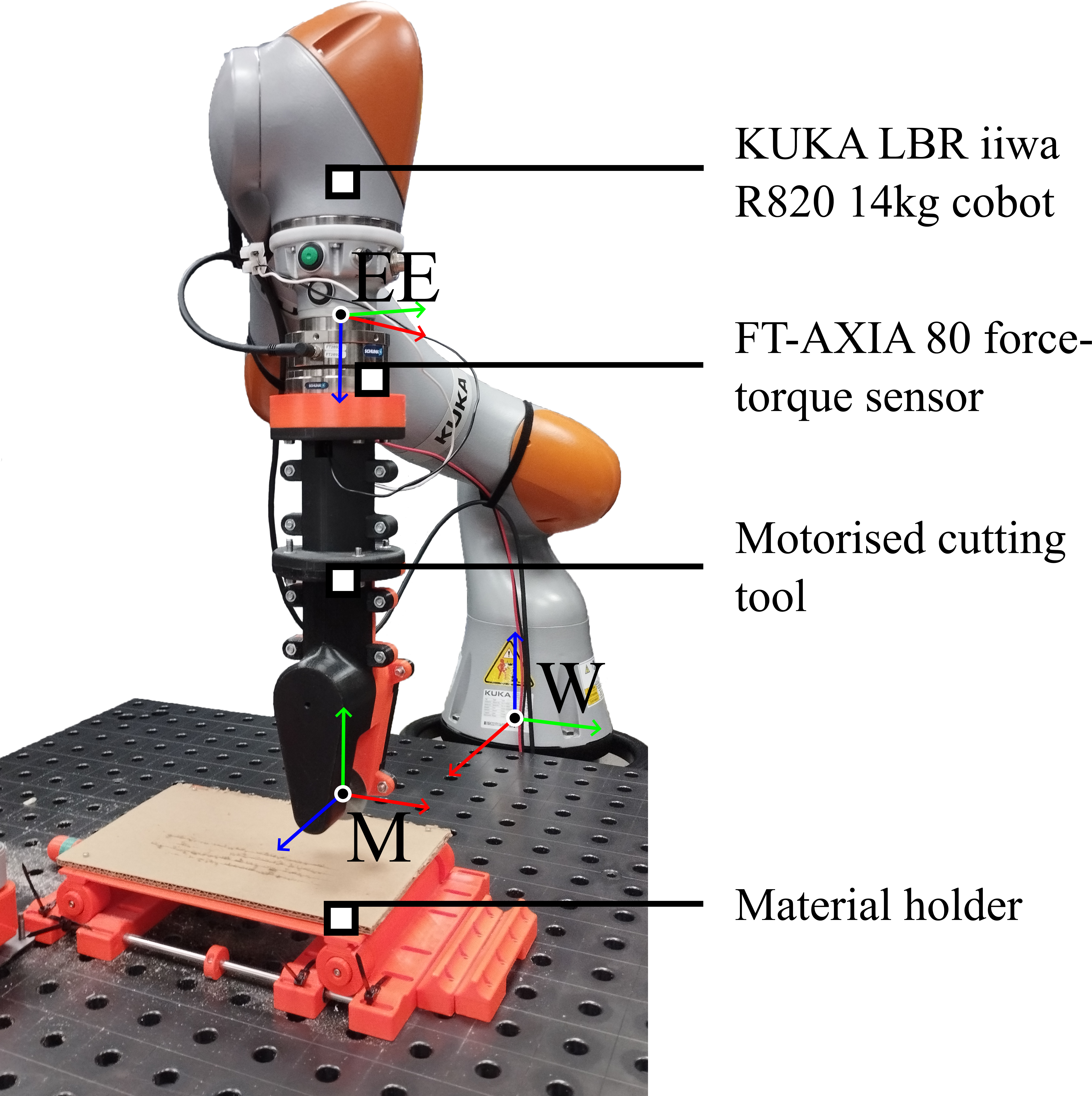}
    \caption{Overview of the experimental setup used for real world cutting experiments, with tool reference frame $\mathcal{M}$, end-effector $\mathcal{EE}$ and world frame $\mathcal{W}$ shown.}
    \label{fig:Method-RealWorld-Setup}
\end{figure}

The real world setup consists of a KUKA LBR \textit{iiwa} R820 14kg collaborative robot equipped with a wrist mounted motorised slitting saw tool. Although the \textit{iiwa} has built-in torque sensing capabilities, we consider the case where a force-torque sensor (FT-AXIA 80) is mounted between the cutter tool and the robot flange which measures the process force. The policy controls the robot via an operational space computed torque tracking controller, with control law:
\begin{equation}
	\vec{\tau} = \mat{J}^{\tpose}\left[\mat{\Lambda}(\vec{q})\left[\mat{K}_{d}(t)\dot{\vec{e}} + \mat{K}_{p}(t)\vec{e}\right] + \mat{\Gamma}+\vec{\mu}\right]
\end{equation}
where $\vec{\tau}$ are the commanded joint torques, $\mat{J}$ the robot Jacobian, and $\mat{\Lambda}$, $\vec{\Gamma}$, $\vec{\mu}$ the robot dynamic parameters corresponding to inertia, Coriolis / centrifugal forces and gravitational forces respectively. The policy outputs correspond to the controller position gain $\mat{K}_{p}$ and translational setpoint adjustment, consisting of a feed rate modification from a nominal feed rate and depth of cut (DoC) offset from the reference trajectory. The damping gain $\mat{K}_{d}$ is computed according to give a critically damped behaviour for the selected position gain. All quantities are referred to in the base frame $W$. The experimental setup is shown in Figure \ref{fig:Method-RealWorld-Setup}, depicting the base ($W$) and model ($M$) frames employed in the modelling approach. 
For each strategy, we consider a single cutting pass of a planar material by conventional milling. The reference tool path is defined with respect to the material surface position, with a reference DoC of $0$mm -- hence, the actual DoC is directly selected via the policy DoC offset. Evaluation was carried out similarly to the simulation as \eqref{eq:RL-Reward-Function}, where MRV was estimated by recording the TCP position during each trial with respect to the ground truth material surface position to obtain a depth of cut, and subsequently MRV by the sum of the swept areas during the task.

\section{Results \& Discussion \label{sec:Results}}

In this section, the performance of the proposed method is evaluated in simulation in the context of the performance of the expert in the source domain (i.e. simulation without GP augmentation) and surrogate target domain (simulation augmented with GP). The performance and behaviour is subsequently compared for the true target domain for a series of real world cutting tasks.

\subsection{Simulation studies}

\begin{figure}
    \centering
    \captionsetup{aboveskip=10pt,belowskip=-18pt}
    \includegraphics[width=0.9\columnwidth]{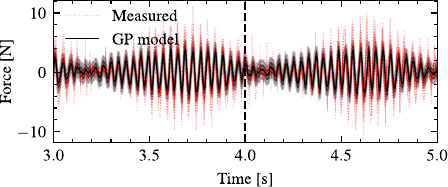}
    \caption{Plot of measured disturbance force in feed (Y) direction from dataset of examples after temporal alignment. The Gaussian process model fit is shown; the shaded areas show 1-$\sigma$ and 2-$\sigma$ standard deviations from the mean respectively. The dashed line shows the transition from training data (left) to extrapolation (right).}
    \label{fig:DTW-Force-Alignment}
\end{figure}

The GP dataset, was constructed from 14 preliminary cutting trials on aluminium and mica, and comprised 35000 samples from 2500-sample windows (5s at 500Hz) per experiment, however, the method is applicable with greater or fewer data. Figure \ref{fig:DTW-Force-Alignment} shows the results of temporal alignment of the measurements overlaid with the fitted GP model. Good agreement between the model predictions and the measured forces is shown over the training data which is similarly shown over the transition from training data to extrapolation.

\begin{figure}[t]
    \centering
    \begin{subfigure}[t]{0.9\columnwidth}
        \centering
        \includegraphics[width=\textwidth]{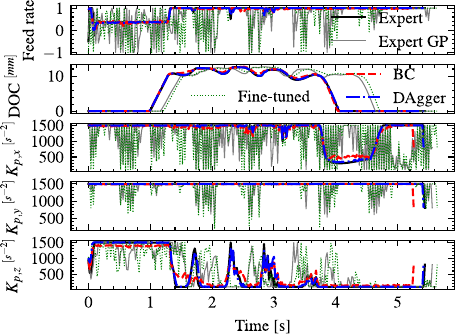}
        \caption{Actions}
        \label{fig:Results-Action-Comparisons-Simulation}
    \end{subfigure}\\
    \begin{subfigure}[t]{0.9\columnwidth}
        \centering
        \includegraphics[width=\textwidth]{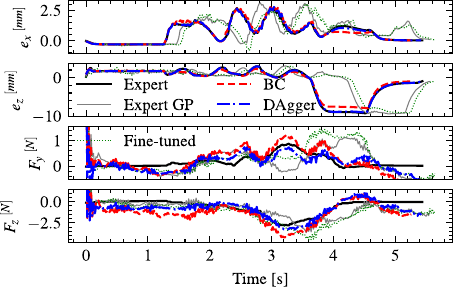}
        \caption{States}
        \label{fig:Results-State-Comparisons-Simulation}
    \end{subfigure}
    \setlength{\belowcaptionskip}{-16pt}
    \caption{Comparison of policy actions and states between source domain expert in source and surrogate target domains (Expert, Expert GP), and surrogate target domain policies using fine-tuning, behavioural cloning (BC) and DAgger imitation learning strategies. States include the path error transverse $e_{x}$ and normal $e_{z}$ to the planned path and forces in the feed $F_{y}$ and normal direction $F_{z}$. Forces are shown with a 50-point (1s) moving average filter.}
    \label{fig:Results-Comparisons-Simulation}
\end{figure}

We first establish a case study for cutting of a planar material from a set of 50 trials with randomly chosen mechanistic constants for the simulation augmented with the learned GP model. We compare the performance of the ``expert'' agent, which is trained in the unaugmented simulation for 32000 episodes and examine the behaviour with respect to the policies trained with the proposed GP + imitation strategy and fine-tuning, with the performance in the base case, i.e. unaugmented simulation as a benchmark. Figure \ref{fig:Results-Comparisons-Simulation} shows the actions adopted by each strategy, consisting of the relative feed rate versus the nominal (0.75m/min), DoC and controller position gain $\vec{K}_{p}$. Correspondingly, the path deviations in the transverse ($e_x$) and normal ($e_z$) directions are shown, along with forces in the feed and normal directions ($F_y$, $F_z$ respectively). Note all quantities are referred to with respect to the robot base frame ($W$, Figure \ref{fig:Method-RealWorld-Setup}) with uniaxial tool feed antiparallel to the y-axis.

From Figure \ref{fig:Results-Action-Comparisons-Simulation} the behaviour of the expert in the source domain can be seen. When transferred directly into the surrogate target domain (``Expert GP''), the policy reacts aggressively to the added disturbance, exhibiting sporadic variations in feed rate and gain selection. In comparison, the surrogate target domain policies trained with both imitation learning strategies closely tracks the source domain expert behaviour. Conversely, while the actions adopted by the original expert policy after fine-tuning differ, particularly for the Z component of position gain, however, the fine-tuned policy remains similarly sensitive to the disturbances. Comparing the states in Figure \ref{fig:Results-State-Comparisons-Simulation} further corroborates the similar performance achieved by both imitation learned policies, with DAgger tracking the expert behaviour more closely than BC. The fine-tuned policy adopts a pattern similar to the expert as directly applied to the surrogate target domain, with a delay of $\sim$0.1s. As an aside, note that with low-pass filtering of the forces, the underlying trend, as seen in the ``expert'' evaluation, is not recovered, even with aggressive cutoff. This also has the effect of introducing delay into the measured signal; for example, a 10th-order Butterworth filter with a cutoff defined at 5Hz has a maximal group delay of $\sim$0.5s, or 25 policy evaluations.

Broadly, for all strategies, similar patterns in actions can be observed during each phase of the cutting task. For example, the policy adopts a high position gain and feed rate broadly around the nominal during approach, followed by transitioning to a high feed rate and low gain in the normal direction (Z) after impact. Furthermore, the DoC remains consistent across all trials. Therefore, the comparison of each strategy implies predominantly \emph{behavioural} benefits of the proposed approach, rather than fundamentally altering the ``decision-making'' of the original expert strategy.

\begin{figure}
    \centering
    \captionsetup{aboveskip=3pt,belowskip=-5pt}
    \includegraphics[width=0.8\columnwidth]{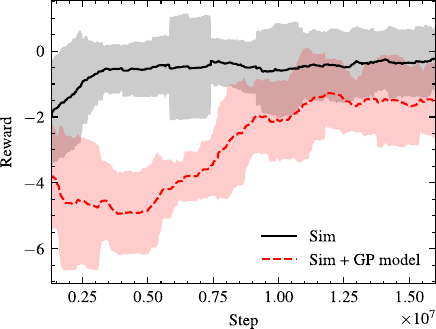}
    \caption{Comparison of training curves between base environment and environment with GP residual force model showing average rewards with 95\% confidence intervals.}
    \label{fig:Results-Training-Curve-Comparison}
\end{figure}

\begin{figure}
    \centering
    \captionsetup{aboveskip=3pt,belowskip=-18pt}
    \includegraphics[width=0.9\columnwidth]{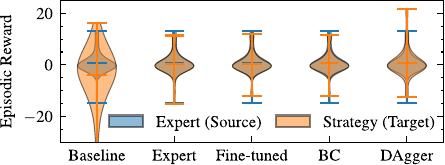}
    \caption{Violin plot of reward distribution between source domain expert policy and target domain strategies -- fixed process parameters for all materials (baseline), unmodified and fine-tuned expert policies, and target policies trained with BC and DAgger imitiation learning approaches.}
    \label{fig:Results-Policy-Dist-Comparison}
\end{figure}

Figure \ref{fig:Results-Training-Curve-Comparison} shows the training curves for expert policies trained from scratch in source and surrogate target domains respectively. The source policy converges rapidly in the first 3M training steps, before a phase of gradual improvement over the remaining 13M steps to a final reward of -0.766. For the surrogate target policy, an initial reduction in performance is recorded, followed by gradual improvement from steps 5.5--12M until convergence to a final reward of -2.24. Besides the advantage of training time reduction (50 episodes vs. 32000, $\sim$0.16\% of training time), the final rewards from the surrogate target expert policy are notably reduced. This follows from the inclusion of the GP model representing a more challenging task for the agent; the force observations in particular are directly related to the reward and encode important information about the interaction which relates to the selection of milling process parameters.

We consider the performance of the original expert policy, the target domain fine tuned policy and policies with each imitation learning strategy (BC, DAgger) in both the original simulation environment and simulation augmented with the learned GP model from real-world trials. We additionally compare the performance with fixed process parameters (Baseline) at the nominal (feed rate 0.75m/min and 1mm DoC) for all materials. In each instance, we compare the episodic rewards over 50 simulation rollouts. From Figure \ref{fig:Results-Policy-Dist-Comparison} it is clear that the overall performance of the expert and all strategies is robust to distributional mismatch between source and target domains. For the fine-tuned, BC and DAgger strategies, the performance is slightly more consistent in the extreme case as indicated by the higher minimum rewards obtained. The greatest deviation is observed for the DAgger trained target policy from the expert for both the original and GP-augmented environments. As DAgger allows for the trajectories to deviate from those obtained using the expert demonstrations alone, it enables the policy to explore and adapt further to the surrogate target domain. It is therefore unsurprising that DAgger demonstrates superior performance in this case.

\subsection{Real world cutting trials}

\begin{figure}
    \centering
    \begin{subfigure}[t]{0.9\columnwidth}
        \centering
        \includegraphics[width=\textwidth]{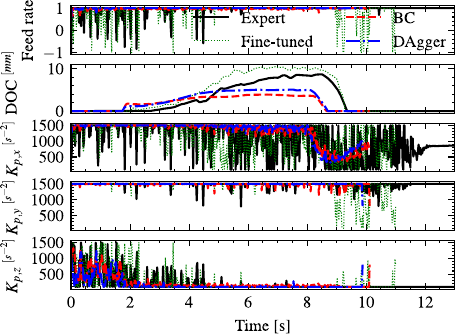}
        \caption{Polyurethane foam}
        \label{fig:Results-Action-Comparisons-Foam}
    \end{subfigure}\\
    \begin{subfigure}[t]{0.9\columnwidth}
        \centering
        \includegraphics[width=\textwidth]{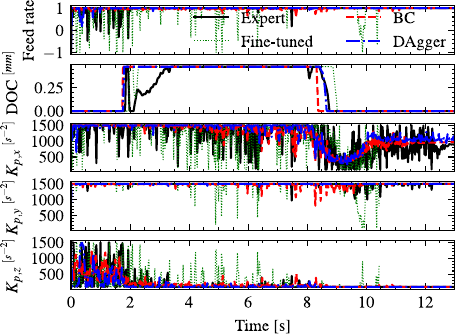}
        \caption{Mica sheet}
        \label{fig:Results-Action-Comparisons-Mica}
    \end{subfigure}
    \captionsetup{aboveskip=3pt,belowskip=-22pt}
    \caption{Comparison of policy actions between source domain expert and surrogate target domain policies using behavioural cloning (BC) and DAgger imitation learning strategies. Actions include the relative feed rate adjustment vs. nominal (0.75m/min), depth of cut (DoC) and controller position gain $\mat{K}_{p}$.}
    \label{fig:Results-Action-Comparisons}
\end{figure}

\begin{figure}[t]
    \centering
    \begin{subfigure}[t]{0.9\columnwidth}
        \centering
        \includegraphics[width=\textwidth]{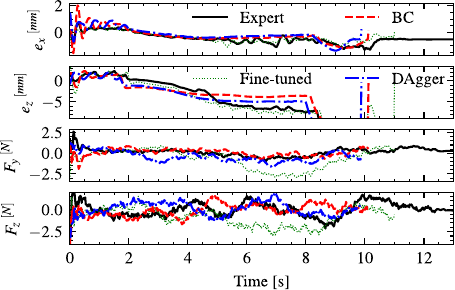}
        \caption{Polyurethane foam}
        \label{fig:Results-State-Comparisons-Foam}
    \end{subfigure}\\
    \begin{subfigure}[t]{0.9\columnwidth}
        \centering
        \includegraphics[width=\textwidth]{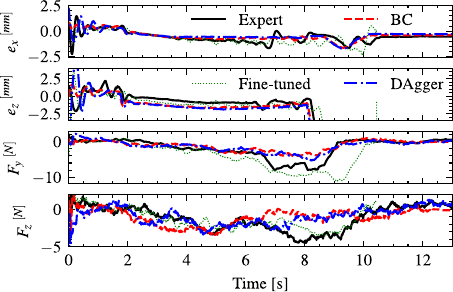}
        \caption{Mica sheet}
        \label{fig:Results-State-Comparisons-Mica}
    \end{subfigure}
    \captionsetup{aboveskip=2pt,belowskip=-22pt}
    \caption{Comparison of path error transverse $e_{x}$ and normal $e_{z}$ to the planned path and forces in the feed $F_{y}$ and normal direction $F_{z}$ between source domain expert, BC and DAgger surrogate target domain policies. Forces are shown with a 50-point (1s) moving average filter.}
    \label{fig:Results-State-Comparisons}
\end{figure}

\begin{table*}
    \centering
    \caption{Comparison of performance metrics for each strategy and material. Metrics comprise completion time $t$, average path deviation $e$, average tool load $f$ (lower better) and material removed volume (MRV, higher better). \textbf{Bold} and \textit{Italic} entries indicate the best performing overall strategy and best performing policy-based method respectively.}
    \resizebox{\textwidth}{!}{%
    \begin{tabular}{r|>{\centering}p{0.35cm}>{\centering}p{0.35cm}>{\centering}p{0.3cm}>{\centering}p{0.6cm}|>{\centering}p{0.35cm}>{\centering}p{0.35cm}>{\centering}p{0.3cm}>{\centering}p{0.6cm}|>{\centering}p{0.35cm}>{\centering}p{0.35cm}>{\centering}p{0.3cm}>{\centering}p{0.6cm}|>{\centering}p{0.35cm}>{\centering}p{0.35cm}>{\centering}p{0.3cm}>{\centering}p{0.6cm}|>{\centering}p{0.35cm}>{\centering}p{0.35cm}>{\centering}p{0.3cm}>{\centering}p{0.6cm}}
    \cline{2-21}
    \multicolumn{1}{c}{} & \multicolumn{4}{c|}{Foam} & \multicolumn{4}{c|}{Plastic} & \multicolumn{4}{c|}{Cardboard} & \multicolumn{4}{c|}{Mica} & \multicolumn{4}{c}{Aluminium}\tabularnewline
    \hline 
    Strategy & $t$ [s] & $e$ [mm] & $f$ [N] & MRV [mm${}^3$] & $t$ [s] & $e$ [mm] & $f$ [N] & MRV [mm${}^3$] & $t$ [s] & $e$ [mm] & $f$ [N] & MRV [mm${}^3$] & $t$ [s] & $e$ [mm] & $f$ [N] & MRV [mm${}^3$] & $t$ [s] & $e$ [mm] & $f$ & MRV [mm${}^3$]\tabularnewline
    \hline
    Baseline   & 19.4 & \textbf{2.29} & 9.03 & \textbf{331} & 20.1 & \textbf{2.29} & \textbf{5.57} & \textbf{72.4} & 20.1 & \textbf{2.38} & 12.1 & \textbf{207} & 19.8 & \textbf{3.67} & 14.1 & 10.0 & 20.1 & \textbf{3.21} & 14.0 & \textbf{17.7} \tabularnewline
    Expert     & 10.6 & 11.6 & 6.54 & 174 & 11.0 & 10.3 & 6.57 & 52.0 & 10.7 & 10.3 & 8.30 & 66.0 & 11.3 & 10.1 & 9.05 & 7.82 & 10.9 & 9.26 & 6.90 & 6.57 \tabularnewline
    Fine-tuned & 10.9 & \textit{5.70} & 8.95 & 202 & 10.5 & \textit{4.41} & 7.49 & 54.0 & 11.0 & \textit{4.35} & 8.32 & 74.0 & 10.5 & \textit{4.46} & 9.84 & \textbf{\textit{15.8}} & 10.3 & \textit{3.96} & 7.59 & \textit{16.0} \tabularnewline
    Re-trained & 44.7 & 8.55 & 10.3 & \textit{259} & 47.6 & 7.44 & 8.61 & 55.3 & 45.4 & 6.83 & 16.0 & 80.2 & 43.8 & 8.47 & 14.4 & 14.2 & 43.7 & 8.12 & 10.4 & 12.3 \tabularnewline
    BC         & 10.6 & 8.26 & 6.23 & 108 & \textbf{\textit{10.3}} & 7.64 & 6.40 & 45.9 & \textbf{\textit{10.4}} & 8.49 & \textbf{\textit{7.44}} & 64.6 & 10.5 & 9.44 & \textbf{\textit{8.42}} & 9.99 & 10.3 & 7.75 & \textbf{\textit{6.58}} & 9.18 \tabularnewline
    DAgger     & \textbf{\textit{10.1}} & 8.95 & \textbf{\textit{6.20}} & 131 & 12.9 & 8.04 & \textit{6.29} & \textit{56.4} & 12.3 & 8.68 & 8.08 & \textit{88.7} & \textbf{\textit{10.3}} & 9.50 & 8.99 & 10.8 & \textbf{\textit{10.2}} & 7.85 & 6.64 & 11.3 \tabularnewline
    \hline 
    \end{tabular}}
    \label{tab:Results-Metrics-Comparison}
    \vspace{-12pt}
\end{table*}

We evaluate the cutting policy for conventional milling of 5 different materials -- high-density polyurethane (PU) foam, corrugated plastic, cardboard, mica, and aluminium -- which exhibit broadly differing mechanical and structural properties. Figure \ref{fig:Results-Action-Comparisons} and Figure \ref{fig:Results-State-Comparisons} show the actions and key states for the original and fine-tuned experts, BC and DAgger trained policies for foam and mica. Emblematic of the source domain expert as applied directly to the real world is a high degree of variability in the agent actions over time, particularly of the controller position gain $\mat{K}_{p}$, corroborating the behaviour observed in simulation. The behaviour of the expert and fine-tuned policies are inconsistent, with path error, depth of cut and force progressively increasing during the cutting task for foam, whereas for the BC and DAgger policies, these are relatively consistent. For the mica cutting task, this is further reflected in the depth of cut, which is improved for the fine-tuned policy. Although all policies saturated at a similar depth of cut, the expert and fine-tuned policies were inferior at regulating the process force, which increased in the feed direction towards the end of the task (7-9s) for the expert, and more extremely for the fine-tuned policy. It is important to note that since the focus of this work is on ``rough'' cutting for disassembly or decommissioning applications, and not on achieving high dimensional tolerances, the expected errors are higher than would be expected for milling, e.g. for a manufacturing application.

\begin{figure}
    \centering
    \includegraphics[width=0.8\columnwidth]{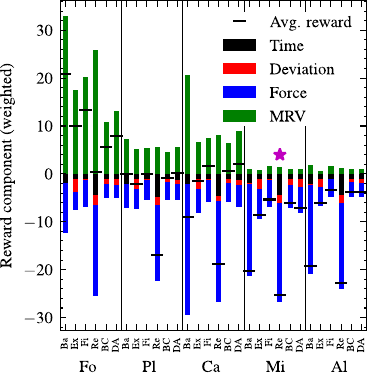}
    \captionsetup{aboveskip=6pt,belowskip=-18pt}
    \caption{Breakdown of average reward and reward components by material and strategy. The performance of each strategy, labelled ``Ba'', ``Ex'', ``Fi'', ``Re'', ``BC'' and ``DA'' respectively are grouped by the material; here abbreviated Foam, Plastic, Cardboard, Mica and Aluminium respectively. Strategies marked with a star were unable to complete the task; the reward is estimated from the single `closest to successful' trial.\vspace{-6pt}}
    \label{fig:Results-Material-Reward-Breakdown}
\end{figure}

Some behavioural characteristics are not preserved from the simulation case study; for example, the selected feed rate during the approach stage of cutting remains nearly at the maximum for all strategies, contrasting the with simulation case. Similarly, the position gain along the normal direction (Z) is increased during the approach phase, however to a lesser extent than in simulation. However, during the cutting phase, these are broadly more similar to the simulation case. As previously established, the force observations encode key information about the interaction, including the discrimination of contact and non-contact states.

The performance metrics - process time, average path deviation, average tool load and MRV - of the expert, fine-tuned, BC and DAgger trained target policies are compared over the average of five cutting trials in Table \ref{tab:Results-Metrics-Comparison}, while the reward components and average reward for all materials is shown in Figure \ref{fig:Results-Material-Reward-Breakdown}. Comparing the expert with direct fine-tuning and with re-training on the surrogate target domain, the fine-tuned policy had consistently lower path deviation and higher MRV, but greater force than the expert in all cases. The re-trained policy similarly had higher MRV and lower path deviation than the expert, albeit much slower task execution and for all materials except aluminium, higher process force than the baseline, making its overall performance inferior to the other policy-based strategies, corroborating the training process in Figure \ref{fig:Results-Training-Curve-Comparison} and emphasising the importance of retaining source domain knowledge for the cutting application. On the other hand, both policies trained with BC and DAgger had consistently lower process force than the other policy-based strategies across all materials. Similarly, path error was consistently lower than the expert for BC and DAgger, although to a lesser extent than the fine-tuned policy. Process time was lowest with the BC and DAgger strategies; DAgger performed best on foam, mica and aluminium, while BC performed best on plastic and cardboard, although reductions in time amongst the policy-based strategies were limited (approx. 3--9\% relative to expert). Adopting the unpaired two-tailed T-test (unequal variances) for overall average rewards, the re-trained policy exhibited inferior performance for all materials ($p<0.001$), whereas fine-tuning yielded significant improvements over the expert for all materials excluding foam ($p<0.05$). Between the imitation learning trained policies and fine-tuned policy, no significant difference in average rewards was found, except for the foam trials ($t(4)=-5, p=0.0047$; $t(4)=-3.1, p=0.018$ for BC, DAgger respectively). This implies fine-tuning and imitation learning are both sufficient for improving target domain policy performance. However, fine-tuning is insufficient to address distributional mismatch in source and target domain actions, leading to inconsistent selection of actions and deviation from behaviour as expected in the source domain. Thus, while the path error is reduced, this comes at the expense of regulating the process force, as shown by the inconsistent force profiles in Figure \ref{fig:Results-State-Comparisons} and metrics in Table \ref{tab:Results-Metrics-Comparison}.

\section{Conclusion\label{sec:Conclusion}}

An imitation-learning based approach for sim-to-real transfer of a robotic cutting policy was proposed. We demonstrate how the residual process dynamics and disturbances can be modelled from a small number of real world trials. We validate the proposed method on a real robot setup, demonstrating the policies transferred to the real world in many cases to have significantly improved performance over the expert as directly transferred from simulation. We also demonstrate the proposed method based on imitation learning outperforms re-training, while performing similarly to fine-tuning even with relatively simple offline imitation learning approaches. Furthermore, the behavioural characteristics of the target domain policies (sensitivity to disturbances) were improved relative to these approaches.

A notable limitation of this work is that while the proposed method can incorporate data from multiple examples, the disturbances are assumed to be sampled from the same underlying process. If the disturbances differ between demonstrations, the alignment between demonstrations will be poor. In this case, the proposed framework could be extended through batched or multi-task GP models. The assumption of a periodic disturbance force furthermore remains a notable limitation. To address these limitations, future work will explore direct synthesis of surrogate real-world data from a minimal unstructured dataset of offline demonstrations.

\bibliographystyle{IEEEtran}
\bibliography{literature}

\begin{thebibliography}{10}
\providecommand{\url}[1]{#1}
\csname url@samestyle\endcsname
\providecommand{\newblock}{\relax}
\providecommand{\bibinfo}[2]{#2}
\providecommand{\BIBentrySTDinterwordspacing}{\spaceskip=0pt\relax}
\providecommand{\BIBentryALTinterwordstretchfactor}{4}
\providecommand{\BIBentryALTinterwordspacing}{\spaceskip=\fontdimen2\font plus
\BIBentryALTinterwordstretchfactor\fontdimen3\font minus
  \fontdimen4\font\relax}
\providecommand{\BIBforeignlanguage}[2]{{%
\expandafter\ifx\csname l@#1\endcsname\relax
\typeout{** WARNING: IEEEtran.bst: No hyphenation pattern has been}%
\typeout{** loaded for the language `#1'. Using the pattern for}%
\typeout{** the default language instead.}%
\else
\language=\csname l@#1\endcsname
\fi
#2}}
\providecommand{\BIBdecl}{\relax}
\BIBdecl

\bibitem{HybridTrajectoryForceLearningAssembly}
Y.~Wang, C.~C. Beltran-Hernandez, W.~Wan, and K.~Harada, ``Hybrid trajectory
  and force learning of complex assembly tasks: A combined learning
  framework,'' \emph{IEEE Access}, vol.~9, pp. 60\,175--60\,186, 2021.

\bibitem{PegInHoleDDPGVIC}
X.~Li, J.~Xiao, W.~Zhao, H.~Liu, and G.~Wang, ``Multiple peg-in-hole compliant
  assembly based on a learning-accelerated deep deterministic policy gradient
  strategy,'' \emph{Industrial Robot: the international journal of robotics
  research and application}, vol.~49, no.~1, pp. 54--64, Jan 2022.

\bibitem{SimToRealDomainRandomisationPushingTask}
X.~B. Peng, M.~Andrychowicz, W.~Zaremba, and P.~Abbeel, ``Sim-to-real transfer
  of robotic control with dynamics randomization,'' in \emph{2018 IEEE
  International Conference on Robotics and Automation (ICRA)}, 2018, pp.
  3803--3810.

\bibitem{LearningRoboticMillingRL}
J.~Hathaway, A.~Rastegarpanah, and R.~Stolkin, ``Learning robotic milling
  strategies based on passive variable operational space interaction control,''
  \emph{IEEE Transactions on Automation Science and Engineering}, pp. 1--14,
  2023.

\bibitem{IdentificationDisturbanceObserver}
K.~Takahei, N.~Suzuki, and E.~Shamoto, ``Identification of the model parameter
  for milling process simulation with sensor-integrated disturbance observer,''
  \emph{Precision Engineering}, vol.~78, pp. 146--162, 2022.

\bibitem{ContinualDeepLearningTimeSeriesModelling}
S.-I. Ao and H.~Fayek, ``Continual deep learning for time series modeling,''
  \emph{Sensors}, vol.~23, no.~16, 2023.

\bibitem{SimToRealRLReview}
E.~Salvato, G.~Fenu, E.~Medvet, and F.~A. Pellegrino, ``Crossing the reality
  gap: A survey on sim-to-real transferability of robot controllers in
  reinforcement learning,'' \emph{IEEE Access}, vol.~9, pp. 153\,171--153\,187,
  2021.

\bibitem{ADATimeDomainAdaptation}
M.~Ragab, E.~Eldele, W.~L. Tan, C.-S. Foo, Z.~Chen, M.~Wu, C.-K. Kwoh, and
  X.~Li, ``Adatime: A benchmarking suite for domain adaptation on time series
  data,'' \emph{ACM Trans. Knowl. Discov. Data}, vol.~17, no.~8, May 2023.

\bibitem{AdversarialDomainAdaptation}
K.~Li, M.~Chen, Y.~Lin, Z.~Li, X.~Jia, and B.~Li, ``A novel adversarial domain
  adaptation transfer learning method for tool wear state prediction,''
  \emph{Knowledge-Based Systems}, vol. 254, p. 109537, 2022.

\bibitem{GenerativeNNBasedDomainAdaptationIncompleteTargetDomain}
C.-B. Chou and C.-H. Lee, ``Generative neural network-based online domain
  adaptation ({GNN-ODA}) approach for incomplete target domain data,''
  \emph{IEEE Transactions on Instrumentation and Measurement}, vol.~72, pp.
  1--10, 2023.

\bibitem{OneShotDomainAdaptiveImitationLearning}
D.~Zhang, W.~Fan, J.~Lloyd, C.~Yang, and N.~F. Lepora, ``One-shot
  domain-adaptive imitation learning via progressive learning applied to
  robotic pouring,'' \emph{IEEE Transactions on Automation Science and
  Engineering}, pp. 1--14, 2022.

\bibitem{UnpairedImage2ImageRL}
P.~M. Scheikl, E.~Tagliabue, B.~Gyenes, M.~Wagner, D.~Dall'Alba, P.~Fiorini,
  and F.~Mathis-Ullrich, ``Sim-to-real transfer for visual reinforcement
  learning of deformable object manipulation for robot-assisted surgery,''
  \emph{IEEE Robotics and Automation Letters}, vol.~8, no.~2, pp. 560--567,
  2023.

\bibitem{DeepSphericalManifoldGPDomainAdaptation}
Y.~Zhang and B.~D. Davison, ``Deep spherical manifold gaussian kernel for
  unsupervised domain adaptation,'' in \emph{2021 IEEE/CVF Conference on
  Computer Vision and Pattern Recognition Workshops (CVPRW)}.\hskip 1em plus
  0.5em minus 0.4em\relax Los Alamitos, CA, USA: IEEE Computer Society, Jun
  2021, pp. 4438--4447.

\bibitem{RLMachiningDeformationControl}
Y.~Zhao, C.~Liu, Z.~Zhiwei, K.~Tang, and D.~He, ``Reinforcement learning method
  for machining deformation control based on meta-invariant feature space,''
  \emph{Visual computing for industry, biomedicine, and art}, vol.~5, p.~27, 11
  2022.

\bibitem{DomainAdaptationRLUnifiedLatentRepresentation}
J.~Xing, T.~Nagata, K.~Chen, X.~Zou, E.~Neftci, and J.~L. Krichmar, ``Domain
  adaptation in reinforcement learning via latent unified state
  representation,'' \emph{CoRR}, vol. abs/2102.05714, 2021.

\bibitem{DomainAdaptationTargetConditionalShift}
K.~Zhang, B.~Scholkopf, K.~Muandet, and Z.~Wang, ``Domain adaptation under
  target and conditional shift,'' in \emph{International Conference on Machine
  Learning}, 2013.

\bibitem{GPDisturbanceObserver}
H.~Jung and S.~Oh, ``Gaussian process and disturbance observer based control
  for disturbance rejection,'' in \emph{2022 IEEE 17th International Conference
  on Advanced Motion Control (AMC)}, 2022, pp. 94--99.

\bibitem{SimToRealNeuralAugmentedSimulation}
F.~Golemo, A.~A. Taiga, A.~Courville, and P.-Y. Oudeyer, ``Sim-to-real transfer
  with neural-augmented robot simulation,'' in \emph{Proceedings of The 2nd
  Conference on Robot Learning}, ser. Proceedings of Machine Learning Research,
  A.~Billard, A.~Dragan, J.~Peters, and J.~Morimoto, Eds., vol.~87.\hskip 1em
  plus 0.5em minus 0.4em\relax PMLR, 29--31 Oct 2018, pp. 817--828.

\bibitem{DeepInverseDynamicModelSim2Real}
P.~F. Christiano, Z.~Shah, I.~Mordatch, J.~Schneider, T.~Blackwell, J.~Tobin,
  P.~Abbeel, and W.~Zaremba, ``Transfer from simulation to real world through
  learning deep inverse dynamics model,'' \emph{CoRR}, vol. abs/1610.03518,
  2016.

\bibitem{SimToRealBasedOnGPandDR}
K.~Wang, J.~Ma, K.~L. Man, K.~Huang, and X.~Huang, ``Sim-to-real transfer with
  domain randomization for maximum power point estimation of photovoltaic
  systems,'' in \emph{2021 IEEE International Conference on Environment and
  Electrical Engineering and 2021 IEEE Industrial and Commercial Power Systems
  Europe (EEEIC / I\&CPS Europe)}, 2021, pp. 1--4.

\bibitem{GPDomainAdaptationMultipleExperts}
S.~Eleftheriadis, O.~Rudovic, M.~P. Deisenroth, and M.~Pantic, ``Gaussian
  process domain experts for model adaptation in facial behavior analysis,'' in
  \emph{2016 IEEE Conference on Computer Vision and Pattern Recognition
  Workshops (CVPRW)}.\hskip 1em plus 0.5em minus 0.4em\relax Los Alamitos, CA,
  USA: IEEE Computer Society, Jul 2016, pp. 1469--1477.

\bibitem{Sim2RealRLWithoutDR}
M.~Kaspar, J.~D. Muñoz~Osorio, and J.~Bock, ``Sim2real transfer for
  reinforcement learning without dynamics randomization,'' in \emph{2020
  IEEE/RSJ International Conference on Intelligent Robots and Systems (IROS)},
  2020, pp. 4383--4388.

\bibitem{OfflineImitationLearningMisspecifiedSim}
S.~Jiang, J.-C. Pang, and Y.~Yu, ``Offline imitation learning with a
  misspecified simulator,'' in \emph{Proceedings of the 34th International
  Conference on Neural Information Processing Systems}, ser. NIPS'20,
  H.~Larochelle, M.~Ranzato, R.~Hadsell, M.~Balcan, and H.~Lin, Eds., vol.~33,
  Red Hook, NY, USA, 2020, pp. 8510--8520.

\bibitem{DTWPackageR}
T.~Giorgino, ``Computing and visualizing dynamic time warping alignments in
  {R}: the dtw package,'' \emph{Journal of statistical Software}, vol.~31, pp.
  1--24, 2009.

\bibitem{TheMachiningOfMetals}
E.~Armarego and R.~Brown, \emph{The Machining of Metals}.\hskip 1em plus 0.5em
  minus 0.4em\relax Prentice-Hall, 1969.

\end{thebibliography}

\end{document}